\documentclass{article} 
\usepackage[dvipsnames]{xcolor}
\usepackage{colm2024_conference}

\usepackage{microtype}
\usepackage{hyperref}
\usepackage{url}
\usepackage{multirow}

\usepackage{xcolor,pifont}
\newcommand*\colourcheck[1]{%
  \expandafter\newcommand\csname #1check\endcsname{\textcolor{#1}{\ding{52}}}%
}
\colourcheck{ForestGreen}

\newcommand{\greentick}{\ForestGreencheck}

\newcommand\Tstrut{\rule{0pt}{2.6ex}}         

\newcommand{\model}{\textsc{Granite-20B-FunctionCalling}}
\newcommand{\granitecode}{\textsc{Granite-20B-Code-Instruct}}

\usepackage{bbm}
\usepackage{todonotes}
\usepackage{multirow}
\usepackage{arydshln}
\usepackage{booktabs}
\usepackage{amsmath}

\usepackage{listings}
\usepackage{xcolor}
\colorlet{punct}{red!60!black}
\definecolor{background}{HTML}{EEEEEE}
\definecolor{delim}{RGB}{20,105,176}
\colorlet{numb}{magenta!60!black}
\usepackage{multirow}
\usepackage{colortbl}
\usepackage[T1]{fontenc}
\lstdefinelanguage{json}{
    basicstyle=\scriptsize\ttfamily,
    numbers=none,
    numberstyle=\scriptsize,
    stepnumber=1,
    numbersep=8pt,
    showstringspaces=false,
    breaklines=true,
    frame=lines,
    backgroundcolor=\color{background},
    literate=
     *{0}{{{\color{numb}0}}}{1}
      {1}{{{\color{numb}1}}}{1}
      {2}{{{\color{numb}2}}}{1}
      {3}{{{\color{numb}3}}}{1}
      {4}{{{\color{numb}4}}}{1}
      {5}{{{\color{numb}5}}}{1}
      {6}{{{\color{numb}6}}}{1}
      {7}{{{\color{numb}7}}}{1}
      {8}{{{\color{numb}8}}}{1}
      {9}{{{\color{numb}9}}}{1}
      {:}{{{\color{punct}{:}}}}{1}
      {,}{{{\color{punct}{,}}}}{1}
      {\{}{{{\color{delim}{\{}}}}{1}
      {\}}{{{\color{delim}{\}}}}}{1}
      {[}{{{\color{delim}{[}}}}{1}
      {]}{{{\color{delim}{]}}}}{1},
}
\lstdefinelanguage{text}{
   basicstyle=\ttfamily,
   columns=fullflexible,
   breakindent=0pt,
   breaklines=true,
   backgroundcolor=\color{gray!15},
   }
\title{Granite-Function Calling Model: Introducing Function Calling Abilities via Multi-task Learning of Granular Tasks}

\usepackage{tikz}

\colmfinalcopy
\begin{document}

\maketitle

\vspace{0.5cm}
\vspace{-2cm}
\begin{center}

\textbf{Ibrahim~Abdelaziz}$^\star$$^\dagger$\quad
\textbf{Kinjal~Basu}$^\star$$^\dagger$\quad
\textbf{Mayank~Agarwal}$^\star$$^\dagger$\quad
\textbf{Sadhana~Kumaravel}\quad
\textbf{Matthew~Stallone}\quad
\textbf{Rameswar~Panda}\quad
\textbf{Yara~Rizk}\quad
\textbf{GP~Bhargav}\quad
\textbf{Maxwell~Crouse}\quad
\textbf{Chulaka~Gunasekara}\quad
\textbf{Shajith~Ikbal}\quad
\textbf{Sachin~Joshi}\quad
\textbf{Hima~Karanam}\quad
\textbf{Vineet~Kumar}\quad
\textbf{Asim~Munawar}\quad
\textbf{Sumit~Neelam}\quad
\textbf{Dinesh~Raghu}\quad
\textbf{Udit~Sharma}\quad
\textbf{Adriana~Meza~Soria}\quad
\textbf{Dheeraj~Sreedhar}\quad
\textbf{Praveen Venkateswaran}\quad
\textbf{Merve~Unuvar}\quad
\textbf{David~Cox}\quad
\textbf{Salim~Roukos}\quad
\textbf{Luis~Lastras}\quad
\textbf{Pavan~Kapanipathi}$^\dagger$\quad

IBM Research \\
\vspace{0.2cm}
$^\star$Equal Contribution \\
$^\dagger$Corresponding Authors \\
\texttt{\{ibrahim.abdelaziz1, kinjal.basu, mayank.agarwal\}@ibm.com, kapanipa@us.ibm.com}

\end{center}

\begin{abstract}

Large language models (LLMs) have recently shown tremendous promise in serving as the backbone to agentic systems, as demonstrated by their performance in multi-faceted, challenging benchmarks like SWE-Bench and Agent-Bench.
However, to realize the true potential of LLMs as autonomous agents, they must learn to identify, call, and interact with external tools and application program interfaces (APIs) to complete complex tasks. These tasks together are termed \textit{function calling}. 
Endowing LLMs with function calling abilities leads to a myriad of advantages, such as access to current and domain-specific information in databases and knowledge sources, and the ability to outsource tasks that can be reliably performed by tools, e.g., a Python interpreter or calculator. 
While there has been significant progress in function calling with LLMs, there is still a dearth of open models that perform on par with proprietary LLMs like GPT, Claude, and Gemini. 
Therefore, in this work, we introduce the \model{}\footnote{The model will be available soon at \url{https://huggingface.co/ibm-granite/}} model under an Apache 2.0 license. The model is trained using a multi-task training approach on seven fundamental tasks encompassed in function calling, those being Nested Function Calling, Function Chaining, Parallel Functions, Function Name Detection, Parameter-Value Pair Detection, Next-Best Function, and Response Generation. We present a comprehensive evaluation on multiple out-of-domain datasets comparing \model{} to more than 15 other best proprietary and open models. 
\model{} provides the best performance among all open models on the Berkeley Function Calling Leaderboard and fourth overall. 
As a result of the diverse tasks and datasets used for training our model, we show that \model{} has better generalizability on multiple tasks in seven different evaluation datasets. 




\end{abstract}
\section{Introduction}


Large language models (LLMs) have garnered significant attention due to their broad applicability to an important set of challenging domains, e.g., programming~\citep{mishra2024granite, roziere2023code}, reasoning~\citep{reid2024gemini,jiang2023mistral}, and multi-modal interaction~\citep{reid2024gemini}. Increasingly, applying these models to solve real-world problems requires them to act as autonomous agents powering intelligent decision-making in specific environments~\citep{yao2022react,xu2023rewoo,yang2024sweagent}\footnote{Auto-GPT:\url{https://github.com/Significant-Gravitas/AutoGPT}}\footnote{BabyAGI:\url{https://github.com/yoheinakajima/babyagi}}. For LLMs to serve as autonomous agents, they must perform accurately on two fundamental capabilities: (a) reasoning and planning, and (b) function calling, which includes identifying, calling, and interacting with tools and APIs in external environments. In this work, we focus on improving LLMs' function calling abilities.

Function calling provides a means for language models to leverage external tools and resources. 
These tools can make available to an LLM specific, up-to-date information that would otherwise be inaccessible (e.g., stored in a dynamic knowledge base) and thus reduce its proclivity for hallucinating responses \citep{Schick2023ToolformerLM}. This is particularly crucial in enterprise use cases where a significant portion of relevant data is stored in a structured format accessible only via storage engines. In addition to knowledge access, function calling can allow an LLM to outsource tasks that are out of scope for a generalized language model. Most commonly, these tasks involve compute-heavy operations, e.g., program execution \citep{shinn2023reflexion}, numerical calculation, or retrieval~\citep{Schick2023ToolformerLM}, and are otherwise a frequent source of LLM hallucinations~\citep{li2023chain}. 

The importance of function calling has spurred the development of several recent 
data generation efforts for fine-tuning~\citep{basu2024apiblend, guo2024api, qin2023toolllm, berkeley-function-calling-leaderboard, tang2023toolalpaca} and evaluation of models~\citep{li2023apibank, muennighoff-etal-2023-crosslingual}. Typically, however, the fine-tuned models from datasets like ToolLLM~\citep{qin2023toolllm}, ToolAlpaca~\citep{tang2023toolalpaca}, and Gorilla~\citep{patil2023gorilla} underperform in one (or more) of three key dimensions: (a) \textbf{Generalizability:} While the datasets are generated using diverse sets of APIs (e.g., ToolLLama uses RapidAPIs~\footnote{\url{https://rapidapi.com/hub}}, ToolAlpaca uses public APIs\footnote{\url{https://github.com/public-apis/public-apis}}, and Gorilla uses TensorFlow Hub, PyTorch Hub, and Hugging Face Hub), work from~\citep{basu2024apiblend} has shown that models trained on these datasets have difficulty generalizing to out-of-domain datasets. (b) \textbf{Granular tasks:} Function calling, as an umbrella term, can encompass multiple granular sub-tasks such as function-name detection, slot filling\footnote{\textit{Slot}, \textit{parameter}, and \textit{argument} are used interchangeably.} or parameter-value pair detection, and detecting the ordered sequence of functions needed to be called. Existing models trained to perform function calling lack the ability to handle these granular tasks independently, and hence, perform poorly on such sub-tasks. (c) \textbf{Openness:} The best performing models are proprietary and the ones that have open licenses (e.g., Gorilla~\citep{patil2023gorilla}) are trained using data generated from OpenAI models.  

\begin{figure}[t]
\begin{center}
\includegraphics[width=\textwidth]{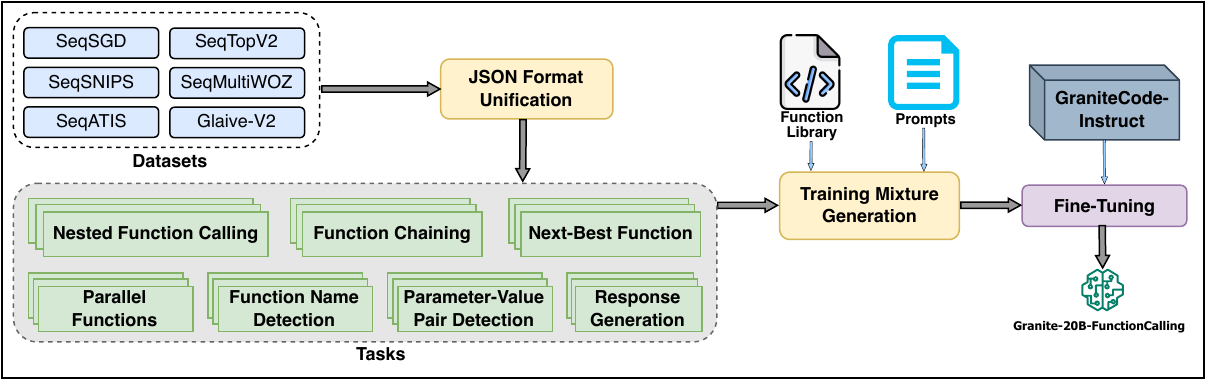}
\caption{Step-by-step building process of \model{}}
\label{fig:fc-model-data-flow}
\end{center}
\end{figure}

\begin{figure}[t]
\begin{center}
\includegraphics[width=0.9\textwidth]{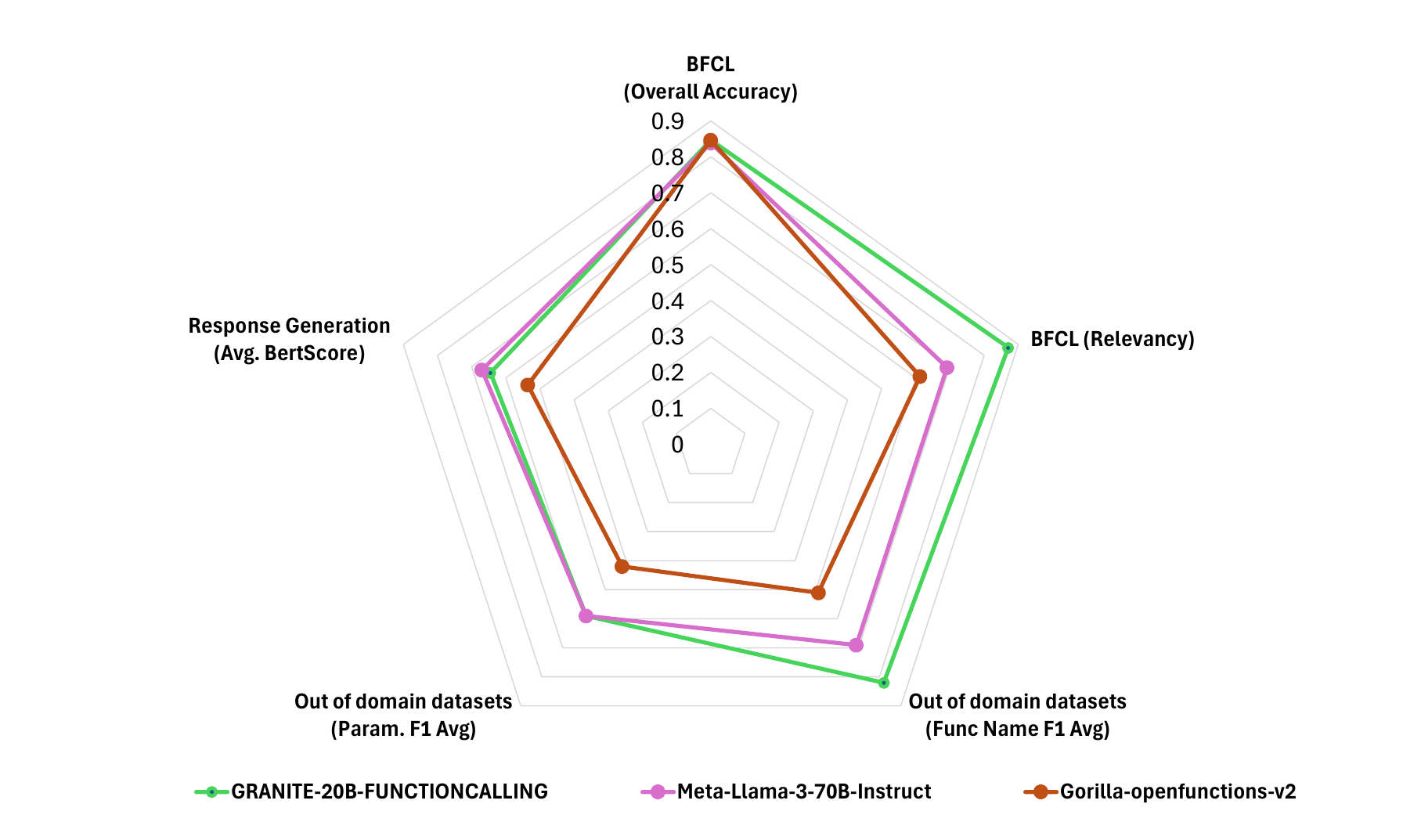}
\caption{Evaluation of \model{} against the best \textit{open} function calling models (according to BFCL)}
\label{fig:radar}
\end{center}
\end{figure}

To address the aforementioned limitations, in this work, we focus on introducing function-calling abilities to models with an inherent focus on granular tasks. Figure \ref{fig:fc-model-data-flow} shows an overview of how \model{} was trained. The datasets used for training are API-Blend~\citep{basu2024apiblend} that include tasks such as function name detection, slot filling, parallel functions, multiple functions, sequencing\footnote{\textit{Sequencing} and \textit{chaining} are used interchangeably.}, and calling APIs\footnote{ \textit{Function} and \textit{API} are used interchangeably.} using multiple programming languages. 
We build upon Granite code models by instruction tuning them for function calling using the datasets for granular tasks with a multi-task learning approach. Granite code models are trained on data that is license permissible following IBM’s AI Ethics principles for trustworthy enterprise usage~\citep{mishra2024granite}. Being part of the Granite family, we release \model{} under Apache 2.0 license. Finally, in this work, we perform a comprehensive evaluation of the open and proprietary models using Berkeley Function Calling Leaderboard (BFCL), four Function Calling Academic Benchmarks, and Response Generation Benchmark from API-Bank \citep{li2023apibank} to evaluate the generalizability of function-calling models. \model{} is on par with the best open model on BFCL and fourth overall.  Furthermore, compared to other models based on the out-of-domain datasets, \model{} shows significant generalizability. 
Figure \ref{fig:radar} shows how \model{} compares to the top two open models (according to BFCL) on various tasks where despite only having 20B parameters, it performs as well or better than Meta-Llama-3-70B-Instruct which has 70B parameters.

\section{Related Work}

\subsection{Instruction Tuning}

Our work is an instantiation of \textit{instruction tuning} \citep{wei2021finetuned}, a fine-tuning method that improves an LLM's ability to solve natural language tasks \citep{mishra2022cross,wang-etal-2023-self-instruct}. It involves taking a large collection of NLP datasets, reformulating those datasets into a set of instruction-following tasks, and then fine-tuning an LLM on the modified data. While the earliest versions of instruction tuning straightforwardly combined large datasets together, the most recent iterations use more sophisticated mixtures of tasks to achieve the best results \citep{li2024synthetic, sudalairaj2024lab}. Our work draws largely upon API-Blend~\citep{basu2024apiblend} and API Pack~\citep{guo2024api}, two recently introduced instruction-tuning datasets specifically focused on tasks related to APIs, e.g., slot filling and API intent detection.

Instruction-tuned models often significantly outperform their base models on a wide range of tasks, particularly in the zero-shot setting \citep{ouyang2022training,muennighoff-etal-2023-crosslingual,chung2022scaling}. Further improvement has been observed through alignment of the model after instruction-fine-tuning, e.g., TP-LLaMA~\citep{chen2024advancing} uses Direct Preference Optimization~\citep{rafailov2023direct} in addition to fine-tuning.

There is also growing interest in developing smaller models that match or surpass the accuracy of larger proprietary LLMs in function calling tasks~\citep{chen2024octopus}. These compact models are crucial in the increasing prevalence of on-device LLMs, enabling efficient and effective performance on local devices.


\subsection{Function Calling by LLMs}

Function calling augmentation has broadened the scope of problems addressable by LLMs to include those that cannot be solved with internal knowledge alone. For instance, prior work has demonstrated the use of API-enhanced LLMs to solve problems requiring up-to-date information retrieval \citep{Schick2023ToolformerLM}, intricate mathematical calculations \citep{he2023solving, patel-etal-2021-nlp}, internet use \citep{komeili2022internet,gur2023real}, task orchestration \citep{jain2024smartflow}, and even programming abilities \citep{gao2023pal}.

Multiple strategies have been proposed for how best to enable LLM function calling. One line of prior research has investigated the design of elaborate prompting approaches, best exemplified by the popular ReACT prompting framework \citep{yao2022react}. Such prompting methods can vary in their design, with some works optimizing for cost \citep{xu2023rewoo}, raw performance \citep{shinn2023reflexion,yang2023mm}, or a blend of both \citep{crouse2023formally}. More relevant to our approach are methods that train models to directly output function calls \citep{tang2023toolalpaca,qin2023toolllm}. Typically, these works will use some form of self-supervision to enable scaling to the breadth of domains required for general-purpose function use \citep{Schick2023ToolformerLM,parisi2022talm,yang2024gpt4tools}.


Recently, many language models with function-calling capabilities have been introduced. They broadly fall into two categories: pre-trained models which are capable of function-calling \citep{reid2024gemini, codegemma_2024, cohereforai-c4ai-command-r, llama-3, jiang2023mistral}, and models fine-tuned specifically for function-calling \citep{qin2023toolllm, tang2023toolalpaca, meetkai-functionary-7b-v1.4, patil2023gorilla, nousresearch2023b, nexusraven}. While the pre-trained models enable function-calling using a combination of supervised and preference fine-tuning, details of the datasets used to train models for these tasks are not generally available. On the other hand, specialized function-calling models mostly rely on synthetic data generated from proprietary state-of-the-art models. Models like Gorilla \citep{patil2023gorilla}, ToolLlama \citep{qin2023toolllm}, ToolAlpaca \citep{tang2023toolalpaca}, and the NousResearch Hermes series of models \citep{nousresearch2023b} utilize GPT-4 or ChatGPT to generate synthetic instruction tuning datasets and fine-tune a base model such as the Llama or Mistral model for function-calling tasks. The NexusRaven models \citep{nexusraven} are one of the few open-source models that focus on building function-calling models for commercial purposes by avoiding using proprietary models for synthetic data generation.

In section \ref{sec_exp}, we compare our model to the above models and show that \model{} provides the best or comparable performance amongst all open models across multiple tasks.

\section{Multi-Task Training Data}
\label{sec_training_data}
In this section, we describe our detailed approach to fine-tune the
\granitecode{} model with multi-task data related to functions to build \model{}, a robust model designed for function-calling. We use API-BLEND~\citep{basu2024apiblend}, a diverse corpora of multiple API datasets for training LLMs. It consists of five datasets with a total of about 160K training examples: SeqSGD, SeqSNIPS, SeqTopV2, SeqATIS, and SeqMultiWOZ. 

A key contribution to the process of building \model{} is multi-task training, where we reuse the same data in different formats with distinct instructions for different function-calling related tasks. We have identified six underlying sub-tasks for function calling and divided them into two broad categories based on the difficulty levels: (A) \textit{High-Level Function Calling Tasks} which are complex tasks for an LLM and typically handle multiple functions; and (B) \textit{Low-Level Function Calling Tasks} which are simpler tasks for an LLM and relate to either function names or only parameter-value pairs. We have included \textit{``Response Generation''} as the seventh task in our training data since producing natural language responses is one of the fundamental goals of an LLM. Table \ref{table:data-task-map} demonstrates the task-wise mapping of each dataset. Below, we describe each task in detail.

\begin{table}[]
\scriptsize
\begin{tabular}{l|ccc|ccc|c}
\toprule
\multicolumn{1}{c|}{}                                    & \multicolumn{3}{c|}{\textbf{High-Level Function Calling Tasks}}                                                                                                                                                                                                                                                                                     & \multicolumn{3}{c|}{\textbf{Low-Level Function Calling Tasks}}                                                                                                                                                                                                                                                                                                  & \multicolumn{1}{c}{}                                                                                          \\
\midrule
\multicolumn{1}{l|}{\multirow{-2}{*}{\textbf{Datasets}}} & \multicolumn{1}{c}{{\color[HTML]{9A0000} \textbf{\begin{tabular}[c]{@{}c@{}}Nested Func.\\ Calling\end{tabular}}}} & \multicolumn{1}{c}{{\color[HTML]{9A0000} \textbf{\begin{tabular}[c]{@{}c@{}}Func. \\ Chaining\end{tabular}}}} & \multicolumn{1}{c|}{{\color[HTML]{9A0000} \textbf{\begin{tabular}[c]{@{}c@{}}Parallel \\ Func.\end{tabular}}}} & \multicolumn{1}{c}{{\color[HTML]{303498} \textbf{\begin{tabular}[c]{@{}c@{}}Next-Best \\ Func.\end{tabular}}}} & \multicolumn{1}{c}{{\color[HTML]{303498} \textbf{\begin{tabular}[c]{@{}c@{}}Func. Name \\ Detection\end{tabular}}}} & \multicolumn{1}{c|}{{\color[HTML]{303498} \textbf{\begin{tabular}[c]{@{}c@{}}Param-Val \\ Pair Detection\end{tabular}}}} & \multicolumn{1}{c}{\multirow{-2}{*}{\textbf{\begin{tabular}[c]{@{}c@{}}Response \\ Generation\end{tabular}}}} \\ \midrule 
{\color[HTML]{343434} \textbf{SeqSGD}}\Tstrut  & & \greentick & \greentick & \greentick & \greentick & \greentick &  \\ 
{\color[HTML]{343434} \textbf{SeqSNIPS}}    & & \greentick &  \greentick & \greentick & \greentick & \greentick &  \\
{\color[HTML]{343434} \textbf{SeqTopV2}}    & \greentick & \greentick &  & \greentick & \greentick & \greentick &  \\
{\color[HTML]{343434} \textbf{SeqATIS}}    & & \greentick &  \greentick & \greentick & \greentick & \greentick &  \\
{\color[HTML]{343434} \textbf{SeqMultiWOZ}}   & & \greentick & & \greentick & \greentick & \greentick &  \\
{\color[HTML]{343434} \textbf{Glaive-V2}}       & & \greentick & & & & & \greentick        \\
\bottomrule
\end{tabular}
\caption{Training Datasets with Task mapping}
\label{table:data-task-map}
\end{table}




In the rest of the section, we describe how to unify the data of different datasets in the same format for model training and then describe each of these training tasks.

\subsection{Data Unification}
In addition to the datasets in API-BLEND~\citep{basu2024apiblend}  (i.e., SeqSGD, SeqSNIPS, SeqTopV2, SeqATIS, SeqMultiWOZ), we also use Glaive-V2\footnote{\url{https://huggingface.co/datasets/glaiveai/glaive-function-calling-v2}} to prepare the training data mixture for \model{}, where each dataset is multi-purposed for different function calling related tasks with different instructions. These datasets come from different sources and have various function formats that require unification to an identical representation for better usability. 

In this unification process, we convert all the APIs, tools, and functions from the data into a JSON format representation. We choose this format because (a) JSON is a language-independent, human-readable, and widely used data format for code-related tasks; (b) it is easily parsable to insert/extract information; and (c) many web services, APIs, and tools accept JSON objects and generate responses in JSON format. 

In \model{}, we unify the model \textit{output} representation of function calls to the following format:

\begin{lstlisting}[language=json]
{
 "name": "<FUNCTION-NAME>", 
 "arguments": {"<PARAMETER-1>": "VALUE-1", "<PARAMETER-2>": ...}
}
\end{lstlisting}

With a similar JSON representation, we express the functions in the library to be passed as input to the model: 
\begin{lstlisting}[language=json]
{
 "name": "<FUNCTION-NAME>", 
 "description: "<FUNCTION-DESCRIPTION>"
 "arguments": {"<PARAMETER-1>": {"description": "...", ...} ... }
}
\end{lstlisting}

\subsection{High-Level Function Calling Tasks}
In general, these tasks are challenging for LLMs to accomplish since they require the LLM to generate multiple function calls with the parameters and their values. In the following sections, we describe different types of high-level function-calling tasks with examples.

\paragraph{Nested Function Calling} The main characteristic of this task is in the output function sequence, where the current function's output becomes an input to the next function. So, the answer to a user query is a sequence of nested function calls selected from the function library. Furthermore, the parameters of these function calls need to be filled by extracting the values from the user query. The example below lists two functions in the output under \texttt{<function\_call>} tag \footnote{In all of our examples, we use \texttt{<function\_call>} tag to represent a function in JSON form.} and the output of the first function will populate the \texttt{destination} parameter in the second function.

\begin{lstlisting}[language=json]
{
 "input": "What's the typical driving time between Las Vegas and the Grand Canyon?", 
 "output: "<function_call> {
             "name": "get_location", 
             "arguments": 
                {"point_on_map": "the Grand Canyon"}
             } 
           <function_call> {
             "name": "get_estimated_duration", 
             "arguments": 
                {"source": "Las Vegas", 
                 "method_travel": "driving", 
                 "destination": "<function_response>get_location"}
             }"
}
\end{lstlisting}

\paragraph{Function Chaining} In this task, a model needs to call multiple functions in a sequence to answer a user query. However, unlike Nested Function Calling, these functions do not have to be nested. Also, for each function, the parameters whose values are present in the user query must be passed as arguments. The following example demonstrates how three functions must be called in sequence to answer the query.

\begin{lstlisting}[language=json]
{
 "input": "Find a provider in SF. Get the weather in SF on March 3rd. Book an appointment with A businessman's haircut on March 3rd at 18:15.", 
 "output: "<function_call> {
             "name": "find_provider", 
             "arguments": 
                 {"city": "SF"}
             } 
           <function_call> {
             "name": "get_weather", 
             "arguments": 
                 {"city": "SF", 
                  "date": "march 3rd"}
             } 
           <function_call> {
             "name": "book_appointment", 
             "arguments": 
                 {"appointment_date": "march 3rd", 
                  "appointment_time": "18:15", 
                  "stylist_name": "A businessman's haircut"}
             }"
}
\end{lstlisting}

\paragraph{Parallel Functions} Similar to the Function Chaining task, here, the answer to a user query requires the same function to be called multiple times (in parallel). Also, the parameters should be mapped with the values extracted from the user query. In the below example, the \texttt{get\_weather} function has been called twice with different parameter-value pairs to answer the user query. 

\begin{lstlisting}[language=json]
{
 "input": "What is the weather like in the next 3 days in New York and in the next 5 days in Los Angeles?", 
 "output: "<function_call> {
             "name": "get_weather", 
             "arguments": 
                 {"current_location": "New York", 
                  "time_range": "next 3 days"}
             } 
           <function_call> {
             "name": "get_weather", 
             "arguments": 
                 {"current_location": "Los Angeles", 
                  "time_range": "next 5 days"}
             }"
}
\end{lstlisting}

\subsection{Low-Level Function Calling Tasks}
Compared to High-Level tasks, these tasks are simpler because they either request function names (without parameters) or parameter-value pairs for a function. To excel in High-Level function calling tasks, it is crucial for any LLM to master these foundational sub-tasks. In what follows, we demonstrate three types of low-level function-calling tasks.

\paragraph{Next-Best Function} In this task, given the function library along with the user query and the partial function sequence, the models are supposed to select the next most suitable function from the function library. It only requires the model to choose one function name without any parameters. The following example demonstrates how the partial function sequence (provided under \texttt{<|partial\_function\_call\_list|>}) can be passed with the query and the model is supposed to predict the next best function. 

\begin{lstlisting}[language=json]
{
 "input": "Get the dates for Raiders Vs Lions in Oakland. Buy 4 tickets for Raiders Vs Lions in Oakland tomorrow. 
      <|partial_function_call_list|> 
            <function_call> {"name": "get_event_dates", "arguments": {}}", 
 "output: "<function_call> {"name": "buy_event_tickets", "arguments": {}}"
}
\end{lstlisting}

\paragraph{Function Name Detection} This task expects the model to produce only the sequence of function names (without parameters) from the function library that are required to answer the user query. This task closely resembles Function Chaining (a High-Level task), with the sole distinction being it does not necessitate the model to populate the function’s arguments. Below is an example of this task.

\begin{lstlisting}[language=json]
{
 "input": "Find the hotel named autumn house. Book a hotel named Autumn House for 2 people for 4 days on Sunday. Find a restaurant in south with name taj tandoori and price range expensive.", 
 "output: "<function_call> {"name": "find_hotel", "arguments": {}} 
           <function_call> {"name": "book_hotel", "arguments": {}} 
           <function_call> {"name": "find_restaurant", "arguments": {}}"
}
\end{lstlisting}

\paragraph{Parameter-Value Pair Detection} In this task, when provided with a user query or a user-agent conversation along with a list of parameters and their descriptions, the model must identify all the parameters for which the values are present in the query or conversation. See an example of this task below. To keep the data format consistent and JSON parsable, we use the function name - \textit{``dummy''} for all the data under this task. 

\begin{lstlisting}[language=json]
{
 "input": "Get a luxury ride to Greentree Apartments with 1 seat.", 
 "output: "<function_call> {
             "name": "dummy", 
             "arguments": 
                 {"destination": "Greentree Apartments", 
                  "number_of_seats": "1", 
                  "ride_type": "Luxury"}
             }"
}
\end{lstlisting}

\subsection{Response Generation}
Natural language response generation is a crucial feature of any LLM. In this task, the model must comprehend an ongoing conversation between a user and an AI assistant. Then, it generates a natural language response, answering the most recent user utterance. Such responses are needed to chit-chat with the user, ask clarifying questions, or synthesize a function call's output into a natural language response.

\begin{lstlisting}[language=json]
{
 "USER": "I have some chicken, broccoli, and cheese. What can I make with these?", 
 "AGENT": "<function_call> {
             "name": "search_recipes", 
             "arguments": 
                 {"ingredients": ["chicken", "broccoli", "cheese"]}
             } 
          <function_response> {
             "recipes": [
                 {"name": "Chicken and Broccoli Casserole", 
                 "ingredients": ["chicken", "broccoli", "cheese"], 
                 "instructions": "Preheat oven to 350 degrees F (175 degrees C). In a large bowl, mix together chicken, broccoli, and cheese. Bake in preheated oven for 25 to 30 minutes, or until cheese is melted and bubbly."}, 
                 {"name": "Cheesy Chicken and Broccoli Pasta", 
                 "ingredients": ["chicken", "broccoli", "cheese"], 
                 "instructions": "Cook pasta according to package instructions. In a large skillet, cook chicken until no longer pink. Add broccoli and cheese, cook until cheese is melted."}]
             }"
 "ASSISTANT": "Here are a couple of recipes you can make with chicken, broccoli, and cheese:..."
}
\end{lstlisting}
\section{Instruct Tuning}

\subsection{Training Data Mixture Creation} 

After generating the data for various tasks, the next step is to create a training data mixture including all the data. We programmatically generate the mixture of data by following a weighted configuration for datasets and tasks. Following is an \textit{example} of the weighted configuration, where the total mixture samples will be divided between Function Chaining and Next-Best Function in a 3:5 ratio. Within the Function Chaining portion, the allocation is split between SeqSGD and Glaive-V2 in a 2:3 ratio. Similarly, the Next-Best Function chunk will be divided in a 2:1 ratio between SeqTopV2 and SeqSNIPS. 

\begin{lstlisting}[language=json]
[{
    "instruction_name": "Function Chaining",
    "datasets": {
        "SeqSGD": 2,
        "Glaive-V2": 3
    },
    "weight": 3
},
{
    "instruction_name": "Next-Best Function",
    "datasets": {
        "SeqTopV2": 2,
        "SeqSNIPS": 1
    },
    "weight": 5
}]
\end{lstlisting}


\begin{table}[h]
  \scriptsize
  \centering
  \begin{tabular}{l p{9cm}}
  \toprule
\textbf{Task} & \textbf{Instruction}  \\\midrule
\textbf{Nested Function Calling} \\\textbf{Function Chaining} \\ \textbf{Parallel Functions} &
\begin{lstlisting}[language=text, aboveskip=-22pt, belowskip=2pt]
SYSTEM: You are a helpful assistant with access to the following function calls. Your task is to produce a sequence of function calls necessary to generate response to the user utterance. Use the following function calls as required.\n<|function_call_library|>\n{API_SPEC_INSTRUCTION}\n\nUSER: {QUERY}\nASSISTANT: 
\end{lstlisting}
\\
\textbf{Next-Best Function} & 
\begin{lstlisting}[language=text, aboveskip=-4pt, belowskip=2pt]
SYSTEM: You are a helpful assistant with access to the following function calls. Your task is to produce the next function call necessary to generate response to the user utterance given the partial function list. Use the following function calls as required and return only function "name" with empty "arguments" dictionary in your response. Once all the necessary functions are called, please return "<|endoftext|>".\n<|function_call_library|>\n{API_SPEC_INSTRUCTION}\n\nUSER: {QUERY}\nASSISTANT:  
\end{lstlisting}
\\
\textbf{Function Name Detection} & 
\begin{lstlisting}[language=text, aboveskip=-4pt, belowskip=2pt]
SYSTEM: You are a helpful assistant with access to the following function calls. Your task is to produce a sequence of function calls necessary to generate response to the user utterance. Use the following function calls as required and return only function "name" with empty "arguments" dictionary in your response. If no function is relevant, please return "<no_function_call>" followed by "<|endoftext|>".\n<|function_call_library|>\n{API_SPEC_INSTRUCTION}\n\nUSER: {QUERY}\nASSISTANT: 
\end{lstlisting}
\\
\textbf{Parameter-Value Pair Detection} & 
\begin{lstlisting}[language=text, aboveskip=-4pt, belowskip=2pt]
SYSTEM: You are a helpful assistant with access to the following function calls. Your task is to find all the necessary arguments and their values from the user utterance to generate response. Use the following function calls as required and fill only the arguments whose values are present in the user utterance.\n<|function_call_library|>\n{API_SPEC_INSTRUCTION}\n\nUSER: {QUERY}\nASSISTANT: 
\end{lstlisting}
\\
\textbf{Response Generation} & 
\begin{lstlisting}[language=text, aboveskip=-4pt, belowskip=2pt]
SYSTEM: You are a helpful assistant with access to the following function calls. Your task is to understand the given conversation with function calls and responses and generate natural language response as the ASSISTANT to continue the conversation. You may use the following function calls to understand how to respond to the user query.\n<|function_call_library|>\n{API_SPEC_INSTRUCTION}\n\n{CONV}\nASSISTANT: 
\end{lstlisting}
\\
  \bottomrule
  \end{tabular}
  \caption{Task specific instructions}
  \label{table:prompt}
\end{table}

Also, in this step, the training data is embedded with the instructions. These instructions are based on the tasks associated with the data. Table \ref{table:prompt} showcases all the instructions (task-wise) we have used in our training. The ``\verb$<|function_call_library|>$'' tag has been used for the function library demonstrated in the prompt with the placeholder named - \verb${API_SPEC_INSTRUCTION}$. As the name suggests, the \verb${QUERY}$ and \verb${CONVERSATION}$ serve as placeholders for user queries or a user-agent conversation, respectively.

\subsection{Training}
\model{} is instruct-tuned version of \granitecode{} \citep{mishra2024granite}\footnote{\url{https://huggingface.co/ibm-granite/granite-20b-code-instruct}}. For training data, we created a mixture of 142K examples spanning all the tasks' datasets discussed above. We then trained our model using QLoRA fine-tuning \citep{dettmers2023qlora} based on our multi-task training mixture discussed above. In particular, we trained \model{}  a QLoRA rank of 8, alpha of 32 and a dropout of 0.1. We also used a learning rate of 5e-5 and ApexFusedAdam as our optimizer with a linear learning rate scheduler. Training was done using a single node of 8 A100\_80GB GPUs with 800GB of RAM for a total of 3 epochs.


\section{Experimental Setup and Evaluation}
\label{sec_exp}
In the section below, we detail our extensive evaluation on various evaluation datasets and public leaderboard. We provide a comprehensive comparison of our \model{}, open sourced with Apache 2.0 license, to other open and proprietary function calling models. 

\subsection{Datasets}
The evaluation datasets and leaderboards for function calling are gaining a lot of traction in the recent past. In particular, to evaluate the models' generalizability, we evaluated \model{} on a variety of function calling benchmarks, all of which are out-of-domain evaluation for our model. It is worth noting that some of these datasets; e.g. ToolAlpaca and ToolLLM, have training data releases. However, we \textit{did not use} any of these benchmarks to train \model{} and we only used the datasets in \ref{table:data-task-map}. \footnote{We could not verify whether some (or all) of the out-of-domain datasets were used in other models' training sets.} Table~\ref{tab:outofdomaindatasets} depicts the details of the evaluation datasets we used. We list the details of each of these evaluation datasets below.



\begin{table}[]
\centering
\resizebox{0.8\textwidth}{!}{%
\begin{tabular}{lcll}
\toprule
Dataset                               & Test Instances & Testing tasks & Metrics \\
\midrule

BFCL &1,700&Function Calling& AST, Execution Accuracy\\
 &&Relevancy& Accuracy\\
ToolLLM &491&Function Calling& Func. matching (F1)\\
RestGPT &157&Function Calling& Func. matching (F1)\\

API-Bank &473&Function Calling& Func. and Param. matching (F1) \\& 478 & 
 Response Generation&  BERTscore, ROUGE, BLEU\\
ToolBench &214&Function Calling& Func. and Param. matching (F1)\\
ToolAlpaca &100&Function Calling& Func. and Param. matching (F1)\\
NexusRaven &318&Function Calling& Func. and Param. matching (F1)\\
\bottomrule
\end{tabular}
}
\caption{Evaluation Datasets}
\label{tab:outofdomaindatasets}
\end{table}


(1) \textbf{Berkeley Function-Calling Leaderboard (BFCL)~\footnote{\url{BFCL: https://gorilla.cs.berkeley.edu/blogs/8_berkeley_function_calling_leaderboard.html}} }
is a comprehensive function calling leaderboard that includes a dataset of over 1,700 instances. The leaderboard evaluates tasks that include (a) Simple Function, Multiple Function, Parallel Function, and Parallel Multiple Function for Python Language; and (b) for non-Python, they evaluate function relevance detection, REST API, JavaScript, and Java. \\
(2) \textbf{ToolBench~\citep{xu2023tool}} is a subset of the data in ToolBench (as released by the authors) focused on HomeSearch and Booking domains. \\
(3) \textbf{ToolLLM}~\citep{qin2023toolllm}\footnote{ToolLLM also calls their benchmark ToolBench. To disambiguate, in this paper we use the term ToolLLM to refer to their benchmark dataset.} is synthetically generated using ChatGPT. The approach uses an initial collection of 16,000 APIs from RapidAPI\footnote{\url{https://rapidapi.com/}} for synthetic data generation. The evaluation is done on the three test sets categorized based on complexity; G1 -- single-tool, G2 -- intra-category multi-tool, G3 -- intra-collection multi-tool. \\
(4) \textbf{RestGPT} \cite{song2023restgpt} is a function calling dataset that has 157 test examples with 85 APIs from Spotify and TMDB. This dataset focuses only on testing model's ability to detect function names. \\
(5) \textbf{API-Bank}~\citep{li2023apibank} has 314 tool-use dialogues with 753 API calls to assess LLMs' capabilities in planning, retrieving, and calling APIs. \\
(6) \textbf{ToolAlpaca} \citep{tang2023toolalpaca} is a synthetic data generation approach that has both training and evaluation benchmarks. It contains 271 tool-use instances spanning 50 distinct categories. Similar to \cite{nexusraven}, we used the simulated part of  ToolAlpaca which has a total of 100 test examples. \\
(7) \textbf{NexusRaven API Evaluation}\footnote{\url{https://huggingface.co/datasets/Nexusflow/NexusRaven_API_evaluation}} is another function calling dataset with 318 test examples covering a total of 65 different APIs. 

\subsection{Evaluation Metrics}
Below, we define the metrics we adopted for specific tasks in function calling. 

\textbf{BFCL Metrics\footnote{\url{https://gorilla.cs.berkeley.edu/blogs/8\_berkeley\_function\_calling\_leaderboard.html\#metrics}}: } BFCL evaluates multiple tasks using the following four metrics. \\
(1) \textit{AST summary} compares the abstract syntax tree of the function output to the ground truth and the function definition. It captures the correctness of the functions called, their parameters (required or not), and the parameter types.  \\
(2) \textit{Execution Summary} compares the execution output from generated and ground-truth function calls. This metric is used to evaluate REST APIs and non-REST data samples. \\
(3) \textit{Relevance} evaluates the model's ability to detect no function calls when the given list of functions is irrelevant to the user query. This inversely captures the hallucination rate of models. \\
(4) \textit{Overall Accuracy} is the weighted average of all individual data splits in BFCL. 

The same metrics described above cannot be used for our out-of-domain datasets because of missing information, varied formats, and response generation task. For example, ToolLLM datasets has missing arguments, ToolAlpaca has missing argument types, and API-Bank has response generation task. Therefore, we use the following metrics to evaluate the models on other datasets:

\textbf{F1 measure}: Based on \cite{basu2024apiblend}, we opted for standard metrics like precision, recall, and F1 scores which focus on exactly matching API and parameters' names. The reason behind this is that APIs are very specific and unless everything (e.g., name, parameters, input/output format, etc.) matches the API specifications, executing such APIs will not be possible. We report F1 for matching function names as well as parameter names and values. 

\textbf{Longest Common Subsequence (LCS) and Exact match}: We also used LCS from \citep{basu2024apiblend} to capture the overlap between the gold and predicted sequences of APIs. This allows us to compute models' ability to predict APIs in the correct sequence as required by the user. Similarly, exact match score ~\citep{basu2024apiblend} checks if all APIs are predicted by the model and are in the same order. 

\textbf{BERTScore, ROUGE-L and BLEU}: We follow the evaluation in API-Bank~\citep{li2023apibank}, a dialog dataset that also evaluates model responses based on language generation metrics such as Rouge-L~\citep{lin2004rouge}, BertScore~\citep{zhang2019bertscore}, and BLEU~\citep{papineni2002bleu}. 

\textbf{Hallucination Rate}: We compute the hallucination rate as the number of samples where the model predicted an API not provided in the function library.

\subsection{Evaluation Results}
Tables~\ref{tab:bfcb-eval},~\ref{tab:ood-fn-eval},~\ref{tab:ood-full-eval},~\ref{tab:apibank-eval}, and Figure~\ref{fig:ood-hallucination-rates} depicts an extensive evaluation of \model{} model in comparison to other state of the art function calling models. In order to detail this evaluation and analyses, below we categorize the results into (a) Berkeley Function Calling Leaderboard Evaluation, and (b) Function calling academic benchmarks.

\begin{table}[]
\resizebox{\textwidth}{!}{%
\begin{tabular}{lllcccc}
\toprule
Model                               & Organization  & License               & AST Summary & Exec. Summary & Relevance & Overall Acc. \\
\midrule
Claude-3.5-Sonnet-20240620 (Prompt)         & Anthropic        & Proprietary           & 91.31       & 89.50        & 85.42     & 90.00          \\
GPT-4-0125-Preview (Prompt)         & OpenAI        & Proprietary           & 91.22       & 88.10         & 70.42     & 88.00          \\
Gemini-1.5-Pro-Preview-0514 (FC)    & Google        & Proprietary           & 87.92       & 83.32        & 89.58     & 86.35       \\
\textbf{\model{}}         & IBM           & Apache 2.0            & 84.11       & 86.50        & 87.08      & 84.71       \\
Gorilla-OpenFunctions-v2 (FC)       & Gorilla   & Apache 2.0            & 89.38       & 81.55        & 61.25     & 84.71       \\
Meta-Llama-3-70B-Instruct (Prompt)  & Meta          & MetaLlama 3          & 87.74       & 85.32        & 69.17     & 83.88       \\
FireFunction-v2                     & Fireworks   & Apache 2.0  & 86.44 & 80.26 & 56.67 & 81.88 \\
Mistral-Medium-2312 (Prompt)        & Mistral AI    & Proprietary           & 83.76       & 73.47        & 88.33     & 81.35       \\
Functionary-Medium-v2.4 (FC)        & MeetKai       & MIT                   & 85.61       & 75.71        & 74.17     & 80.47       \\
Command-R-Plus (Prompt) (Opt.) & Cohere & cc-by-nc-4.0          & 83.60        & 86.74        & 54.17     & 80.35       \\
Functionary-Small-v2.4 (FC)         & MeetKai       & MIT                   & 83.55       & 76.31        & 67.92     & 79.94       \\
Mistral-large-2402 (FC Auto)        & Mistral AI    & Proprietary           & 64.73       & 60.01        & 84.17     & 68.76       \\
Nexusflow-Raven-v2 (FC)              & Nexusflow    & Apache 2.0 & 65.19      & 73.89        & 57.5     & 67.35     \\
DBRX-Instruct (Prompt)              & Databricks    & Databricks & 66.62       & 74.92        & 55.83     & 65.88      \\
Snowflake-arctic-Instruct (Prompt)             & Snowflake    & Apache 2.0 & 61.09       & 80.04        & 59.58     & 65.18      \\
\bottomrule
\end{tabular}
}
\caption{Berkeley Function Calling Benchmark: Top 15 models by Overall Accuracy (as of 06/25/2024). All evaluations are done in a \textit{zero-shot} manner.}
\label{tab:bfcb-eval}
\end{table}

\subsubsection{BFCL Leaderboard Evaluation Results} Table~\ref{tab:bfcb-eval} shows that \model{} is ranked fourth on the overall accuracy metric among the top 15 models on BFCL and is highest among models with open licenses\footnote{We have picked the best performing version of each model. For example, Gemini-1.5-Pro-Preview-0514 (FC) and Gemini-1.5-Pro-Preview-0409 (FC) are both part of the leaderboard but for our evaluation, we consider the best of Gemini-1.5-Pro.}. While it is tied with the Gorilla~\citep{patil2023gorilla} model, it is important to note that the latter was finetuned on data that are (a) generated from ChatGPT, and (b) similar data to the test set and hasn't generalized well to other datasets as shown in Table~\ref{tab:ood-fn-eval} and Figure~\ref{fig:ood-hallucination-rates}. In the context of model sizes, \model{} is one of the smallest models in the list. Specifically, the ones better than \model{} in the ranking are all significantly larger in size. 

For the BFCL evaluation dataset, we highlight concerns in certain categories, particularly the Java, JavaScript, and REST API evaluations. We are concerned with how the Java and JavaScript categories evaluate a function-calling model's capabilities to follow language-specific syntax, for instance how objects are instantiated and called in Java and JavaScript utilizing language-specific context and norms. For the REST API category, we observed significant brittleness in the evaluation due to issues with API availability and API call limits.

\subsubsection{Function Calling Academic Benchmarks}
Tables~\ref{tab:ood-fn-eval} and ~\ref{tab:ood-full-eval} focus on evaluating the models' performance on Function Matching using F1-measure, LCS, and Exact Match. In this experiment, we reuse the model handlers from the BFCL code base, including the optimized prompts for each model. However, since the Cohere Command-R-v01 and Mistral-Instruct-v0.3 handlers available in BFCL use the REST API interface for inference, we reimplement handlers for these models, utilizing local models using prompts suggested by the respective model developers for function calling.

\textbf{Function Name Detection:} On ToolLLM datasets (G1, G2, and G3) and RestGPT, \model{} performs the best on detecting function names given a natural language utterance with \textbf{8\%} better F1 score than the next best function calling model, as shown in Table~\ref{tab:ood-fn-eval}. Since these datasets have multiple functions in sequence, we also compute sequencing metrics; exact score and LCS. On this front, \model{} model also outperforms other function calling models by \textbf{7\%} on LCS and \textbf{11\%} on Exact Match scores.

\begin{table}[]
\resizebox{\textwidth}{!}{%
\begin{tabular}{l|ccc|ccc|ccc|ccc||ccc}
\toprule
\multirow{2}{*}{}             & \multicolumn{3}{c|}{ToolLLM-G1}                                                   & \multicolumn{3}{c|}{ToolLLM-G2}                                                   & \multicolumn{3}{c|}{ToolLLM-G3}  &  \multicolumn{3}{c||}{RestGPT}   &  \multicolumn{3}{c}{Average}     \\
\midrule
                              & \begin{tabular}[c]{@{}c@{}}Func. \\Match \end{tabular} & LCS & \begin{tabular}[c]{@{}c@{}}Exact \\Score\end{tabular} & \begin{tabular}[c]{@{}c@{}}Func. \\Match \end{tabular} & LCS & \begin{tabular}[c]{@{}c@{}}Exact \\Score  \end{tabular} & \begin{tabular}[c]{@{}c@{}}Func. \\Match\end{tabular} & LCS & \begin{tabular}[c]{@{}c@{}}Exact \\Score  \end{tabular}  & \begin{tabular}[c]{@{}c@{}}Func. \\Match \end{tabular} & LCS & \begin{tabular}[c]{@{}c@{}}Exact \\Score  \end{tabular}  & \begin{tabular}[c]{@{}c@{}}Func. \\Match\end{tabular} & LCS & \begin{tabular}[c]{@{}c@{}}Exact \\Score  \end{tabular} \\
                              \midrule
Functionary-small-v2.4 (7B)   & 0.00 & 0.00 & 0.00 & 0.00 & 0.00 & 0.00 & 0.00 & 0.00 & 0.00 & 0.29 & 0.30 & 0.06 & 0.07 & 0.07 & 0.02 \\
Gorilla-openfunctions-v2 (7B) & 0.59 & 0.59 & 0.28 & 0.48 & 0.48 & 0.22 & 0.51 & 0.52 & 0.24 & 0.21 & 0.21 & 0.01 & 0.44 & 0.45 & 0.19 \\
Hermes-2-Pro-Mistral (7B)     & 0.00 & 0.00 & 0.00 & 0.00 & 0.00 & 0.00 & 0.00 & 0.00 & 0.00 & 0.03 & 0.03 & 0.01 & 0.01 & 0.01 & 0.00 \\
Mistral-Instruct-v0.3 (7B)    & 0.49 & 0.49 & 0.26 & 0.51 & 0.49 & 0.30 & 0.36 & 0.33 & 0.13 & 0.36 & 0.37 & 0.08 & 0.43 & 0.42 & 0.19  \\
CodeGemma-Instruct (7B)       & 0.59 & 0.59 & 0.21 & 0.53 & 0.53 & 0.13 & 0.52 & 0.54 & 0.16 & 0.22 & 0.23 & 0.02 & 0.46 & 0.47 & 0.13  \\
Nexusflow-Raven-v2 (13B)              & 0.65 & 0.65 & 0.39 & 0.73 & 0.72 & 0.43 & 0.68 & 0.66 & 0.27 & 0.39 & 0.41 & 0.06 & 0.61 & 0.61 & 0.28 \\
C4AI-Command-R-v01 (35B)      & 0.65 & 0.64 & 0.39 & 0.73 & 0.71 & 0.45 & 0.69 & 0.68 & 0.23 & \textbf{0.59} & \textbf{0.60} & \textbf{0.22} & \underline{0.66} & \underline{0.66} & \underline{0.32}  \\
Meta-Llama-3-70B-Instruct (70B)        & 0.61 & 0.61 & 0.31 & 0.59 & 0.58 & 0.21 & 0.65 & 0.64 & 0.23 & 0.22 & 0.22 & 0.01 & 0.52 & 0.51 & 0.19\\
\textbf{\model{}}                    & \textbf{0.86} & \textbf{0.85} & \textbf{0.63} & \textbf{0.84} & \textbf{0.82} & \textbf{0.58} & \textbf{0.76} & \textbf{0.73} & \textbf{0.35} & 0.51 & 0.52 & 0.15 & \textbf{0.74} & \textbf{0.73} & \textbf{0.43}\\		
\bottomrule
\end{tabular}
}
\caption{Function Calling Academic Benchmarks: Function Name Detection. Best performance is highlighted in \textbf{bold}, second best is \underline{underlined}. All evaluations are done in a \textit{zero-shot} manner. }
\label{tab:ood-fn-eval}
\end{table}


\textbf{Full Function Calling:} Table~\ref{tab:ood-full-eval} reports on the models' performance on the API-Bank, ToolBench, and ToolAlpaca datasets that are out-of-domain and evaluated in a zero-shot manner. No single model outperforms all other models across datasets. Note that datasets like ToolAlpaca and API-Bank come with training data split which we never used for training \model{}, but could not guarantee that the other models were not trained with it too. Averaging out the F1 scores across datasets shows that \model{} achieves an F1 score of 0.87 when predicting the function name; second best by 0.01 to Cohere's Command-R (a 35B model) which provides an F1 score of 0.88. When predicting the arguments, \model{} average F1 score lags behind the best model (Cohere's Command-R) by 0.03; 0.62 vs. 0.59.

\begin{figure}
    \centering
    \includegraphics[width=0.8\textwidth]{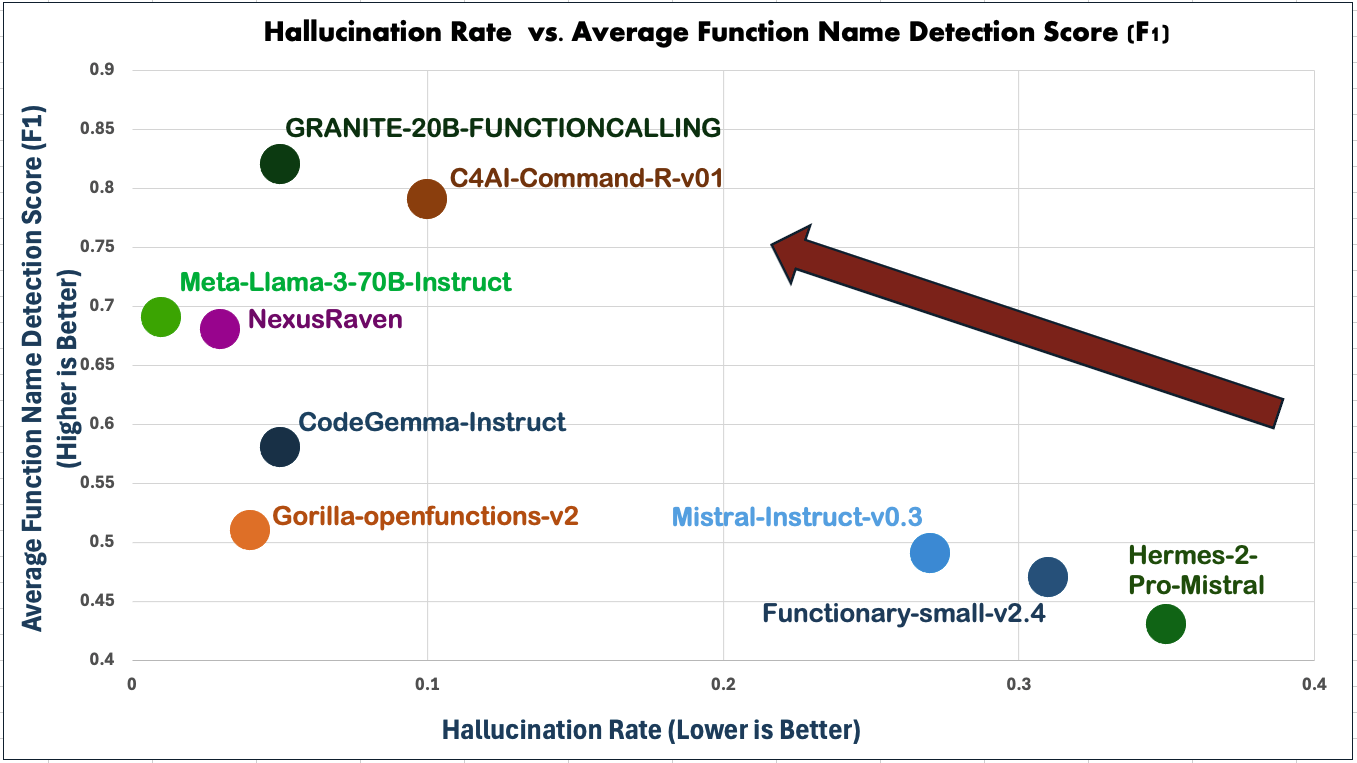}
    \caption{Performance vs. Hallucination rates for Out-of-Domain Function Calling}
    \label{fig:ood-hallucination-rates}
\end{figure}

\textbf{Function Name Hallucination:} Hallucinations have been a major drawback of large language models. In the context of calling and executing APIs, hallucinations can have adverse consequences. In Figure~\ref{fig:ood-hallucination-rates}, we compare the models' Function Name Detection Scores (average F1) over all the datasets (except BFCL, which uses AST-based metrics) and their hallucination rates. Ideally, we want models to have high performance and low hallucination rates placing them in the top left corner of the plot. \model{} has the highest performance with less than 0.1 hallucination rate.

\begin{table}[t]
\resizebox{\textwidth}{!}{%
\begin{tabular}{lcccccc|cc}
\toprule
\multirow{2}{*}{}                      & \multicolumn{6}{c|}{Func-Name+Args Det. (F1 Func-Name | F1 Args)}         & \multicolumn{2}{c}{F1 Average} \\
\midrule
                                       & \begin{tabular}[c]{@{}c@{}}API-Bank\\L-1\end{tabular}  & \begin{tabular}[c]{@{}c@{}}API-Bank\\L-2\end{tabular} & \begin{tabular}[c]{@{}c@{}}ToolBench\\HS\end{tabular} & \begin{tabular}[c]{@{}c@{}}ToolBench\\B\end{tabular} & \begin{tabular}[c]{@{}c@{}}Tool-Alpaca\end{tabular} & \begin{tabular}[c]{@{}c@{}}Nexus\\Raven\end{tabular} & \begin{tabular}[c]{@{}c@{}}Func\\Name\end{tabular}  & Args   \\
                                       \midrule
Functionary-small-v2.4 (7B)            & 0.78 | 0.70 & 0.54 | 0.45 & 0.73 | 0.68 & 0.65 | 0.33 & 0.88 | \textbf{0.47} & 0.82 | 0.64 & 0.73 & 0.55  \\
Gorilla-openfunctions-v2 (7B)          & 0.43 | 0.41 & 0.12 | 0.12 & 0.86 | 0.69 & 0.41 | 0.27 & 0.69 | 0.39 & 0.81 | 0.65 & 0.55 & 0.42  \\
Hermes-2-Pro-Mistral (7B)              & \textbf{0.93} | \textbf{0.77} & 0.54 | 0.25 & 0.51 | 0.40 & 0.56 | 0.26 & 0.80 | 0.26 & 0.90 | 0.63 & 0.71 & 0.43  \\
Mistral-Instruct-v0.3 (7B)             & 0.79 | 0.69 & 0.69 | 0.46 & 0.60 | 0.47 & 0.04 | 0.16 & 0.33 | 0.33 & 0.71 | 0.54 & 0.53 & 0.44  \\
CodeGemma-Instruct (7B)                & 0.77 | 0.57 & 0.59 | 0.38 & 0.65 | 0.50 & 0.54 | 0.22 & 0.59 | 0.31 & 0.84 | 0.68 & 0.66 & 0.44  \\
Nexusflow-Raven-v2 (13B)                       & 0.51 | 0.42 & 0.28 | 0.22 & \textbf{0.92} | 0.65 & 0.89 | 0.35 & 0.85 | 0.37 & 0.92 | \textbf{0.75} & 0.73 & 0.46  \\
C4AI-Command-R-v01 (35B)               & \textbf{0.93} | 0.76 & 0.77 | 0.54 & 0.85 | 0.77 & 0.88 | 0.49 & \textbf{0.90} | 0.42 & \textbf{0.93} | 0.71 & \textbf{0.88} & \textbf{0.62}  \\
Meta-Llama-3-70B-Instruct (70B)                 & 0.85 | 0.67 & 0.69 | 0.52 & 0.91 | \textbf{0.86} & \textbf{0.91} | \textbf{0.56} & 0.78 | 0.43 & 0.70 | 0.52 & 0.81 & \underline{0.59}  \\
\textbf{\model{}} & 0.91 | 0.71 & \textbf{0.83} |\textbf{ 0.60} & 0.87 | 0.71 & 0.82 | 0.36 & 0.89 | 0.44 & 0.92 | 0.72 & \underline{0.87} & \underline{0.59} \\
\bottomrule
\end{tabular}
}
\caption{Function Calling Academic Benchmarks: Full Function Calling. Best performance is highlighted in \textbf{bold}, second best is \underline{underlined}. All evaluations are done in a \textit{zero-shot} manner.}
\label{tab:ood-full-eval}
\end{table}

\begin{table}[t]
\centering
\resizebox{0.9\textwidth}{!}{%
\begin{tabular}{lccc|ccc}
\toprule
                                  & \multicolumn{3}{c|}{API-Bank-Response-Level 1} & \multicolumn{3}{c}{API-Bank-Response-Level 2} \\
                                \midrule
Models                            & BertScore  & Rouge-L  & BLEU   & BertScore  & Rouge-L  & BLEU   \\
\midrule
Functionary-small-v2.4 (7B)       & 0.34       & 0.23     & 0.05  &  0.35       & 0.23     & 0.05  \\
Gorilla-openfunctions-v2 (7B)     & 0.56       & 0.33     & 0.32  &   0.51       & 0.26     & 0.25  \\
Hermes-2-Pro-Mistral (7B)         & 0.45       & 0.18     & 0.09  & 0.42       & 0.14     & 0.06  \\
Mistral-Instruct-v0.3 (7B)        & 0.52       & 0.29     & 0.22  &  0.46       & 0.20      & 0.14   \\
CodeGemma-Instruct (7B)           & 0.14       & 0.03     & 0.00     &  0.09       & 0.02     & 0.01  \\
Nexusflow-Raven-v2 (13B)                  & 0.41       & 0.16     & 0.11  &  0.38       & 0.11     & 0.06  \\
C4AI-Command-R-v01 (35B)          & 0.39       & 0.15     & 0.07  & 0.39       & 0.15     & 0.06  \\
Meta-Llama-3-70B-Instruct (70B)            & \textbf{0.69}       & \textbf{0.48}     & \textbf{0.47}  &          \textbf{0.65} &	\textbf{0.40} &	\textbf{0.40}   \\
\textbf{\model{}} & \underline{0.68}       & \underline{0.47}     & \textbf{0.47}  &  \underline{0.61}       & \underline{0.36}     & \underline{0.37} \\
\bottomrule
\end{tabular}
}
\caption{API-Bank Response generation dataset evaluation. Results are averaged across each dataset per model. Best performance is highlighted in \textbf{bold}, second best is \underline{underlined}. All evaluations are done in a \textit{zero-shot} manner.}
\label{tab:apibank-eval}
\end{table}

\subsubsection{Response Generation} 
In Table~\ref{tab:apibank-eval}, we  show models' performance on response generation task. We use API-Bank dataset and follow their response generation task evaluation with BertScore, Rouge-L, and BLUE. Meta-Llama-3-70B-Instruct has the best performance across the three metrics with \model{} coming in close second (difference in performance ranged between 1-5\%). Both models significantly outperform all other evaluated models. The gap is larger when we compare \model{} to the ones specifically trained for function calling such as Functionary-small-v2.5 and Gorilla-openfunctions-v2.

\subsubsection{Further Improvements}
We have instruct-tuned the \model{} in such a way that it develops implicit function searching capability from a long list of functions. For example, in out-of-domain evaluation tasks, for ToolAlpaca the model needs to find the Function from a list of 94 functions, similarly, it has access to 15 and 20 functions for ToolBench-B and ToolBench-HS, respectively. Due to this reason, the prompt with all the function libraries increases the context length and \granitecode{} supports up to 8192 context length, so we were not able to add the full signature of each Function in the library. Currently, each function in the library contains a function name, a description, and for each function, we have provided a list of respected arguments with their descriptions. However, to fit the prompt in the max context-length, we had to remove the type, required-fields, and optional-fields values for each argument from the specifications. For further performance exploration, we will assess options to include the entire function specification without truncation including exploiting the benefit of Rotary Position Embedding \citep{su2023roformer} and its innate support for longer context lengths within some of the other Granite models.

\section{Conclusion}
\label{sec:conclusion}
In this paper, we introduced \model{}, a capable function calling open model with Apache 2 license. \model{} is trained using a suite of datasets transformed from semantic parsing, task-oriented dialog, personal assistants and conversational domains. The training setup is a multi-task learning approach where granular tasks in function calling such as function detection, parameter detection, sequencing, and next best function are used for instruction tuning the model. We performed an extensive evaluation of \model{} in comparison to other state-of-the-art function calling models. On multiple out-of-domain datasets, including Berkeley Function Calling Leaderboard, \model{} provides the best performance among the models that have open licenses. Even compared to multiple proprietary models with much larger sizes, \model{} showed on-par and in some cases better performance on multiple datasets and tasks. 

\bibliography{colm2024_conference}
\bibliographystyle{colm2024_conference}


\end{document}